\documentclass[conference]{IEEEtran}
\IEEEoverridecommandlockouts
\usepackage{cite}
\usepackage{amsmath,amssymb,amsfonts}
\usepackage{algorithmic}
\usepackage{graphicx}
\usepackage{textcomp}
\usepackage{xcolor}
\usepackage{multirow}
\usepackage{pifont}
\newcommand{\cmark}{\ding{51}} % ✓
\newcommand{\xmark}{\ding{55}} % ✗

\def\BibTeX{{\rm B\kern-.05em{\sc i\kern-.025em b}\kern-.08em
    T\kern-.1667em\lower.7ex\hbox{E}\kern-.125emX}}

\begin{document}

\title{Prototype-Based Low Altitude UAV Semantic Segmentation}

\author{
	\IEEEauthorblockN{
		Da Zhang\textsuperscript{1}, 
		Junyu Gao\textsuperscript{1}, 
		Zhiyuan Zhao\textsuperscript{2*}
	}
	\IEEEauthorblockA{
		\textsuperscript{1}\textit{School of Artificial Intelligence, Optics and Electronics, Northwestern Polytechnical University} \\
		\textsuperscript{2}\textit{Institute of Artificial Intelligence (TeleAI), China Telecom} \\
		zhangda1018@126.com, \{gjy3035, tuzixini\}@gmail.com
	}
	\thanks{*Corresponding author. This work was supported by the National Natural Science Foundation of China under Grant 62576284 and 62306241, and was supported by grants from the Innovation Foundation for Doctor Dissertation of Northwestern Polytechnical University CX2025109.}
}

\maketitle

\begin{abstract}
Semantic segmentation of low-altitude UAV imagery presents unique challenges due to extreme scale variations, complex object boundaries, and limited computational resources on edge devices. 
Existing transformer-based segmentation methods achieve remarkable performance but incur high computational overhead, while lightweight approaches struggle to capture fine-grained details in high-resolution aerial scenes. 
To address these limitations, we propose PBSeg, an efficient prototype-based segmentation framework tailored for UAV applications. 
PBSeg introduces a novel prototype-based cross-attention (PBCA) that exploits feature redundancy to reduce computational complexity while maintaining segmentation quality. 
The framework incorporates an efficient multi-scale feature extraction module that combines deformable convolutions (DConv) with context-aware modulation (CAM) to capture both local details and global semantics. 
Experiments on two challenging UAV datasets demonstrate the effectiveness of the proposed approach. 
PBSeg achieves 71.86\% mIoU on UAVid and 80.92\% mIoU on UDD6, establishing competitive performance while maintaining computational efficiency. Code is available at https://github.com/zhangda1018/PBSeg.
\end{abstract}


\section{Introduction}

With the rapid advancement of consumer-grade UAV technology, low-altitude aerial imaging has emerged as a transformative tool for urban monitoring and environmental analysis \cite{cheng2024methods}. 
UAV platforms provide unprecedented flexibility in data acquisition, offering high-resolution imagery from oblique and nadir perspectives that traditional satellite-based systems cannot achieve \cite{li2021abcnet}. 
These capabilities have enabled diverse applications ranging from urban planning and infrastructure inspection to disaster response and precision agriculture \cite{li2021multiattention}. 

Among the various computer vision tasks applied to UAV imagery, semantic segmentation has gained considerable attention as a fundamental technique for scene understanding \cite{wang2022novel, zhang2024integrating}. 
Recent years have witnessed substantial progress driven by deep learning architectures. 
Convolutional neural networks such as U-Net \cite{ronneberger2015u} established strong baselines through multi-scale feature extraction and encoder-decoder designs \cite{wang2022unetformer, gao2025combining}. 
More recently, transformer-based approaches including Mask2Former \cite{cheng2022masked} have demonstrated superior performance by leveraging global context modeling and unified frameworks for multiple segmentation tasks.
These methods achieve promising results on benchmark datasets through sophisticated attention mechanisms and end-to-end learning paradigms \cite{wang2021transformer}.

\begin{figure}[t]
	\centering
	\includegraphics[width=\linewidth]{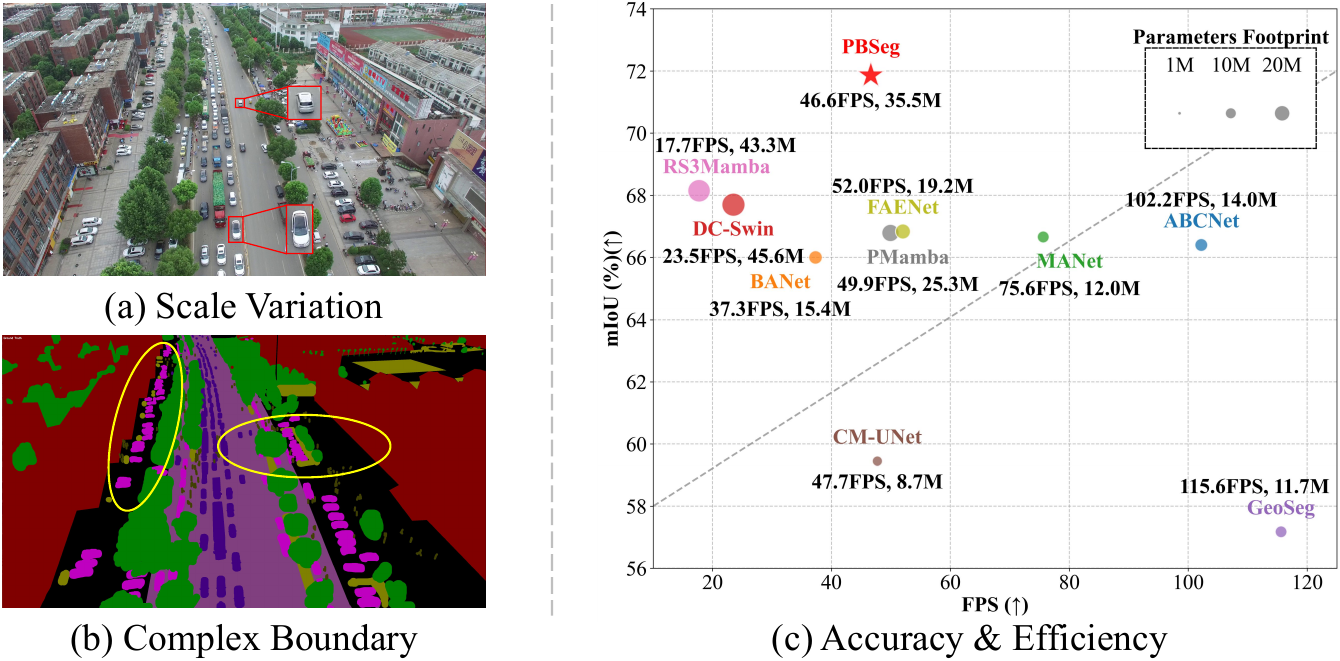}
	\caption{Illustration of the main challenges in low-altitude UAV semantic segmentation. 
		(a) Strong scale variation with small objects embedded in large-scale urban layouts. 
		(b) Complex boundaries and crowded regions make small instances difficult to segment reliably. 
		(c) Trade-off between segmentation accuracy and computational cost on UAVid.
		}
	\label{f1}
\end{figure}

Despite these advances, several challenges (as shown in Fig. \ref{f1}) remain in semantic segmentation of low-altitude UAV imagery \cite{lyu2020uavid}.
Extreme scale variation and complex object boundaries make it difficult to preserve small instances such as vehicles and pedestrians throughout the feature hierarchy. 
In addition, practical deployment is constrained by limited memory and computational resources on onboard processors, which makes many high-capacity models unsuitable for real flights. These limitations necessitate novel architectural designs that balance accuracy with efficiency.

To address these challenges, we propose PBSeg, a prototype-based segmentation framework specifically designed for low-altitude UAV applications. 
PBSeg introduces a prototype-based cross-attention (PBCA) that exploits redundancy in dense feature maps, so that object queries interact only with a compact set of representative prototypes rather than all spatial tokens. 
This design reduces computational complexity while retaining discriminative information for query refinement. 
Furthermore, PBSeg employs an efficient multi-scale pixel decoder that combines deformable convolutions (DConv) with context-aware modulation (CAM) to aggregate local details and global semantics, without relying on transformer layers in the decoder. 
The overall framework aims to achieve a favorable balance between segmentation quality and computational efficiency on resource-constrained UAV platforms.

\begin{figure*}[ht]
	\centering
	\includegraphics[width=\linewidth]{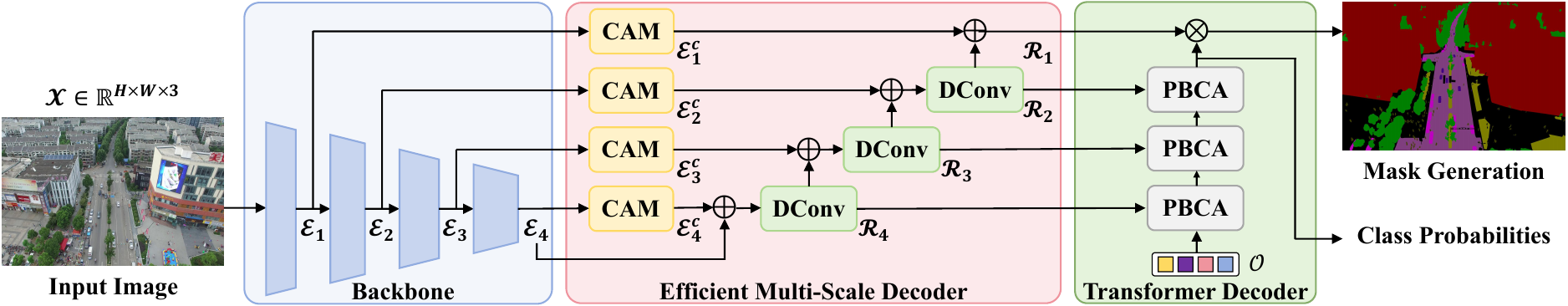}
	\caption{Architecture of PBSeg. The backbone extracts features from the input image; 
the multi-scale decoder upsamples features to recover high-resolution representations; the transformer decoder takes as input a set of learnable object embeddings and the high-resolution features and produces refined queries for inference.
	}
	\label{f2}
\end{figure*}

The main contributions are summarized as follows:
\begin{itemize}
	\item We develop a prototype-based cross-attention that leverages feature redundancy to reduce computational cost while preserving discriminative query representations.
	\item  We design an efficient multi-scale feature pyramid with deformable convolutions and context-aware modulation, which enhances global context modeling in a computationally lightweight manner.
	\item  Experiments on UAVid and UDD6 demonstrate that PBSeg achieves competitive performance compared to SOTA methods while maintaining favorable computational efficiency.
\end{itemize}

\section{Related Work}

\subsection{UAV Semantic Segmentation}

Semantic segmentation of UAV imagery is important for urban monitoring and environmental analysis \cite{cheng2024methods}. Early work mainly adapts generic segmentation networks such as U-Net to aerial scenes \cite{ronneberger2015u}, and EMNet augments Deeplabv3+ with edge feature fusion and multi-level upsampling for UAVid and related datasets to better handle scale variation and boundary complexity \cite{li2023semantic}. Transformer-based frameworks further exploit global context, including the context-aware transformer by Kumar et al.\ \cite{kumar2022semantic} and the composite UAVformer network for urban UAV scenes \cite{yi2023uavformer}. UAV-FAENet and global–local attention networks explicitly target small objects and irregular boundaries by combining frequency-aware modules with attention-based refinement \cite{zhou2025uav, zhang2025uav}. These methods improve accuracy but often rely on heavy backbones or decoder structures, which limits their applicability on resource-constrained UAV platforms \cite{cui2023cm, wang2022unetformer}. PBSeg instead adopts prototype-based attention and a compact convolutional decoder tailored to the efficiency requirements of low-altitude UAV semantic segmentation.

\subsection{Efficient Segmentation Architectures}

Semantic segmentation on embedded or edge hardware requires architectures that balance accuracy and computational cost. Mask-based transformer frameworks such as Mask2Former unify semantic, instance and panoptic segmentation with query-based mask classification \cite{cheng2022masked}, yet multi-head attention and high-resolution pixel decoders still require substantial memory and computation \cite{yu2022k}. To improve efficiency, recent work investigates compact backbones and lightweight decoders \cite{zhang2025kaid, zhang2026boosting}. SegNeXt and SeaFormer use convolutional and axial transformer blocks to provide strong context modeling with lower cost than standard transformer self-attention and achieve favorable accuracy–runtime trade-offs on mobile devices \cite{guo2022segnext, wan2023seaformer}. Hybrid convolutional and transformer designs including LETNet and UNeXt combine lightweight encoders with efficient decoders and achieve real-time performance on driving and remote sensing benchmarks with small parameter budgets \cite{xu2023lightweight, chang2024unext}. These advances show that architectural choices can reduce the cost of dense prediction, but most efficient models target generic ground-level scenes and do not explicitly exploit redundancy in UAV feature representations \cite{wang2024pyramidmamba}. PBSeg introduces prototype-based cross-attention and a compact decoder that jointly address redundancy and multi-scale context in low-altitude UAV imagery.

\section{Method}

\subsection{Framework Overview}

Semantic segmentation aims to assign each pixel to a predefined category. Given an input image $\mathcal{X} \in \mathbb{R}^{H \times W \times 3}$, the task produces $L$ binary segmentation masks $\mathcal{Y} \in \{0,1\}^{L \times H \times W}$ along with class probabilities $\mathcal{P}_j \in \Delta^{C+1}$ for each mask, where $C$ denotes the number of semantic classes and an additional null class accounts for background regions.

PBSeg adopts the mask classification paradigm established by recent unified frameworks (as shown in Fig. \ref{f2}). The architecture comprises three components. A backbone network extracts hierarchical representations $\mathcal{E}_s \in \mathbb{R}^{h_s \times w_s \times d_s}$ at multiple scales $s \in \{1,2,3,4\}$, where spatial dimensions $h_s$ and $w_s$ correspond to the input resolution downsampled by factors $\{4, 8, 16, 32\}$ respectively. An efficient pixel decoder processes these backbone features to generate refined multi-scale representations $\mathcal{R}_s \in \mathbb{R}^{h_s \times w_s \times D}$ with consistent channel dimension $D$ across scales. A transformer decoder refines $L$ learnable object embeddings $\mathcal{O} \in \mathbb{R}^{L \times D}$ through cross-attention with the decoder features $\mathcal{R}_s$ at scales $s \in \{2,3,4\}$, producing updated embeddings $\mathcal{O}' \in \mathbb{R}^{L \times D}$. Final predictions combine the refined embeddings with the highest-resolution decoder output $\mathcal{R}_1$ through matrix multiplication to generate masks, while a classification head applied to $\mathcal{O}'$ predicts semantic categories.

\begin{figure}[ht]
	\centering
	\includegraphics[width=\linewidth]{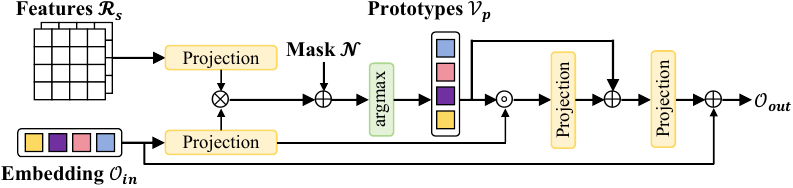}
	\caption{Scheme of the proposed prototype-based cross-attention.
	}
	\label{f3}
\end{figure}

\subsection{Prototype-Based Cross-Attention}

Standard cross-attention in transformer decoders computes interactions between object embeddings and all spatial locations in visual features. 
For high-resolution UAV imagery, this results in quadratic complexity that scales poorly with image size. We observe that pixels belonging to the same semantic region exhibit similar feature representations after sufficient training iterations. This redundancy suggests that refining object embeddings requires only a representative subset of visual tokens rather than exhaustive pixel-wise attention.

Our prototype-based attention mechanism (Fig. \ref{f3}) selects one representative feature per object embedding to reduce computational cost. Given multi-scale features $\mathcal{R}_s$ and object embeddings $\mathcal{O}_{in}$, we first apply linear projections to obtain key representations $\mathcal{V} \in \mathbb{R}^{h_s w_s \times d}$ and query representations $\mathcal{T} \in \mathbb{R}^{L \times d}$ in a shared embedding space of dimension $d$. An affinity matrix $\mathcal{A} \in \mathbb{R}^{h_s w_s \times L}$ captures correspondence between spatial locations and object embeddings through
\begin{equation}
	\mathcal{A} = \mathcal{V} \mathcal{T}^{\top}.
\end{equation}

For each object embedding, we identify the spatial location with maximum affinity as its prototype. To focus selection on foreground regions and avoid background interference, we apply a binary attention mask $\mathcal{N}$ derived from predictions in the previous decoder layer. The mask assigns zero weight to pixels classified as foreground in the prior iteration and negative infinity elsewhere, ensuring prototype selection concentrates on relevant object regions. Formally, prototype indices $\mathcal{J}$ and corresponding features $\mathcal{V}_p \in \mathbb{R}^{L \times d}$ are obtained via
\begin{equation}
	\mathcal{J} = \arg\max_{h_s w_s}(\mathcal{A} + \mathcal{N}), \quad \mathcal{V}_p = \mathcal{V}[\mathcal{J}],
\end{equation}
where $\mathcal{V}[\mathcal{J}]$ indexes $\mathcal{V}$ along the spatial dimension according to $\mathcal{J}$. This selection reduces the number of visual tokens from $h_s w_s$ to $L$, substantially decreasing subsequent computational requirements.

After prototype selection establishes one-to-one correspondence between embeddings and visual features, we compute attention through efficient element-wise operations rather than full pairwise dot products. The interaction between $\mathcal{T}$ and $\mathcal{V}_p$ is modeled as
\begin{equation}
	\mathcal{U} = (\mathcal{T} \circ \mathcal{V}_p) \mathcal{W}_1,
\end{equation}
where $\circ$ denotes element-wise multiplication and $\mathcal{W}_1 \in \mathbb{R}^{d \times d}$ is a learnable projection matrix. Channel-wise normalization and scaling with a learnable parameter $\boldsymbol{\beta} \in \mathbb{R}^d$ produces modulation weights that adjust prototype features additively,
\begin{equation}
	\mathcal{Z} = \boldsymbol{\beta} \circ \frac{\mathcal{U}}{\|\mathcal{U}\|_2} + \mathcal{V}_p.
\end{equation}
A final projection $\mathcal{W}_2 \in \mathbb{R}^{d \times D}$ maps the modulated features back to the embedding dimension, followed by a residual connection to preserve input information,
\begin{equation}
	\mathcal{O}_{out} = \mathcal{Z} \mathcal{W}_2 + \mathcal{O}_{in}.
\end{equation}

This design avoids the $\mathcal{O}(h_s w_s \times L)$ complexity of standard attention while maintaining expressive capacity through prototype-guided refinement. The entire process operates in a multi-head configuration where features and embeddings are partitioned along the channel dimension, allowing different heads to capture diverse semantic relationships.

\subsection{Efficient Multi-Scale Decoder}

The pixel decoder constructs a feature pyramid that provides multi-scale representations for the transformer decoder. While recent methods enhance feature pyramids with deformable attention to capture global context and adaptive receptive fields, such designs introduce substantial computational overhead unsuitable for resource-constrained UAV platforms. We employ a fully convolutional architecture that achieves similar benefits through lightweight operations.

Backbone features $\mathcal{E}_s$ contain rich spatial details but lack holistic scene understanding. We apply context-aware modulation to inject global information efficiently at each scale. For features at stage $s$, a projection reduces dimensionality to $D$, yielding $\mathcal{E}'_s \in \mathbb{R}^{h_s \times w_s \times D}$. Global average pooling followed by a two-layer network with $1 \times 1$ convolutions produces a context descriptor $\boldsymbol{\gamma}_s \in \mathbb{R}^{1 \times D}$,
\begin{equation}
	\boldsymbol{\gamma}_s = \operatorname{MLP}(\operatorname{Pool}(\mathcal{E}'_s)).
\end{equation}

Channel-wise modulation applies sigmoid activation to $\boldsymbol{\gamma}_s$ and combines it with the original features through element-wise multiplication and residual addition,
\begin{equation}
	\mathcal{E}^c_s = \mathcal{E}'_s \circ \sigma(\boldsymbol{\gamma}_s) + \mathcal{E}'_s,
\end{equation}
where $\sigma$ denotes the sigmoid function. This operation reweights channels according to global scene statistics, suppressing irrelevant information while amplifying discriminative features.

\begin{table*}[t]
	\caption{Quantitative comparison results of the UAVid dataset}
	\label{tab:uavid_quant}
	\centering
	\scriptsize
	\renewcommand{\arraystretch}{1}
	
	\resizebox{\linewidth}{!}{%
		\begin{tabular}{l|cccccccc|c|c|c}
			\hline
			\multirow{2}{*}{Method} &
			\multicolumn{8}{c|}{Per-Class IoU (\%)} &
			\multirow{2}{*}{mIoU (\%)} &
			\multirow{2}{*}{mF1 (\%)} &
			\multirow{2}{*}{OA (\%)} \\
			\cline{2-9}
			& Building & Road & Tree & LowVeg & MovingCar & StaticCar & Human & Clutter & & & \\
			\hline
			ABCNet \cite{li2021abcnet}           & 85.77 & 80.92 & 77.88 & 62.46 & 74.12 & 53.33 & 29.29 & 66.14 & 66.40 & 78.13 & 85.48 \\
			BANet \cite{wang2021transformer}     & 83.90 & 81.48 & 78.51 & 62.18 & 77.44 & 52.01 & 29.51 & 64.58 & 66.00 & 77.86 & 84.99 \\
			MANet \cite{li2021multiattention}    & 84.42 & 80.17 & 78.75 & 62.39 & 68.73 & 62.73 & 29.40 & 66.17 & 66.66 & 78.50 & 85.19 \\
			DC-Swin \cite{wang2022novel}         & 87.64 & 82.49 & \textbf{79.78} & 64.64 & 76.02 & 53.62 & 29.72 & \textbf{68.43} & 67.70 & 79.01 & 86.58 \\
			GeoSeg \cite{wang2022unetformer}     & 75.38 & 75.36 & 72.11 & 56.95 & 64.57 & 35.89 & 20.03 & 55.86 & 57.18 & 70.42 & 79.75 \\
			CM-UNet \cite{cui2023cm}             & 79.05 & 78.71 & 73.55 & 56.85 & 69.71 & 33.64 & 24.64 & 59.30 & 59.45 & 72.24 & 81.59 \\
			RS3Mamba \cite{ma2024rs}             & 86.54 & 81.83 & 79.03 & 63.10 & 75.79 & 60.42 & 30.36 & 67.22 & 68.15 & 79.51 & 86.01 \\
			PMamba \cite{wang2024pyramidmamba}   & 85.93 & 81.20 & 78.66 & 62.38 & 73.95 & 56.45 & 28.96 & 66.48 & 66.79 & 78.43 & 85.59 \\
			FAENet \cite{zhou2025uav}            & 87.37 & \textbf{83.20} & 79.26 & 63.54 & \textbf{78.02} & 59.53 & 30.98 & 68.25 & 68.84 & 79.97 & 86.48 \\
			\hline
			PBSeg(Ours) & \textbf{92.13} & 80.55 & 77.88 & \textbf{67.38} & 77.61 & \textbf{70.83} & \textbf{42.25} & 66.27 & \textbf{71.86} & \textbf{82.91} & \textbf{88.01} \\
			\hline
		\end{tabular}%
	}
\end{table*}

\begin{table*}[h]
	\caption{Quantitative comparison results on the UDD6 dataset}
	\label{tab:udd6_quant}
	\centering
	\scriptsize
	\renewcommand{\arraystretch}{1}
	
	\resizebox{\linewidth}{!}{%
		\fontsize{4}{5}\selectfont
		\begin{tabular}{l|cccccc|c|c|c}
			\hline
			\multirow{2}{*}{Methods} &
			\multicolumn{6}{c|}{Per-Class IoU (\%)} &
			\multirow{2}{*}{mIoU (\%)} &
			\multirow{2}{*}{mF1 (\%)} &
			\multirow{2}{*}{OA (\%)} \\
			\cline{2-7}
			& Fa\c{c}ade & Road & Vegetation & Vehicle & Roof & Other & & & \\
			\hline
			ABCNet \cite{li2021abcnet}                & 73.10 & 70.05 & \textbf{89.79} & 71.64 & 88.68 & 63.27 & 78.65 & 87.76 & 88.13 \\
			BANet \cite{wang2021transformer}          & 73.72 & 70.59 & 89.52          & 71.75 & 87.77 & 62.96 & 78.67 & 87.83 & 88.03 \\
			MANet \cite{li2021multiattention}         & 72.53 & 68.84 & 89.49          & 70.37 & 87.64 & 62.58 & 77.78 & 87.22 & 87.58 \\
			DC-Swin \cite{wang2022novel}              & {75.50} &  {73.03} & 89.72 & 72.08 & 89.14 & \textbf{64.95} & 79.89 & 88.61 & {88.95} \\
			GeoSeg \cite{wang2022unetformer}      & 55.87 & 58.40 & 87.16          & 48.16 & 76.08 & 50.20 & 65.13 & 78.00 & 80.39 \\
			CM-UNet \cite{cui2023cm}                  & 64.76 & 66.26 & 88.88          & 60.38 & 84.94 & 55.71 & 73.04 & 83.91 & 85.06 \\
			RS3Mamba \cite{ma2024rs}              & 75.46 & 72.14 & 89.68          & 72.01 & 89.57 & 64.44 & 79.77 & 88.52 & 88.87 \\
			PMamba \cite{wang2024pyramidmamba}    & 74.30 & 71.04 & 89.31          & 70.97 & 89.06 & 63.79 & 78.94 & 87.98 & 88.38 \\
			UNetMamba \cite{zhu2024unetmamba}            & 73.52 & 71.17 & 89.33          & 72.71 & 88.78 & 63.22 & 79.10 & 88.10 & 88.19 \\
			UAV-FAENet \cite{zhou2025uav}             & 75.39 & 72.43 & 89.58          &  {73.27} & \textbf{89.89} & 63.78 & \textbf{80.11} & \textbf{88.74} & 88.80 \\
			\hline
			PBSeg (Ours)   &  \textbf{76.67} & \textbf{73.71} & 89.78 & \textbf{75.59} &  88.87 & 62.94  & 80.92  & 88.56  & \textbf{88.97} \\
			\hline
		\end{tabular}%
	}
\end{table*}

To construct the feature pyramid, we aggregate modulated features across scales using deformable convolutions that adapt receptive fields based on input content. Starting from the lowest resolution $s=4$, we compute pyramid features $\mathcal{R}_s$ recursively. At the coarsest scale, global pooling and projection of $\mathcal{E}_4$ provide additional scene context before deformable convolution,
\begin{equation}
	\mathcal{R}_4 = \operatorname{DConv}(\mathcal{E}^c_4 + \operatorname{Proj}(\operatorname{Pool}(\mathcal{E}_4))).
\end{equation}

For intermediate scales $s \in \{2,3\}$, upsampled features from the previous pyramid level are combined with modulated backbone features through deformable convolution,
\begin{equation}
	\mathcal{R}_s = \operatorname{DConv}(\mathcal{E}^c_s + \operatorname{Upsample}(\mathcal{R}_{s+1})).
\end{equation}

At the finest resolution $s=1$, we omit deformable convolution and directly sum upsampled features with $\mathcal{E}^c_1$ to preserve high-frequency details. The resulting pyramid provides features $\mathcal{R}_s$ at scales $s \in \{2,3,4\}$ for cross-attention in the transformer decoder, while $\mathcal{R}_1$ serves as the high-resolution feature map for final mask generation.

% in preamble
% \usepackage{graphicx}
% \usepackage{multirow}

 \section{Experiments}
 
 \subsection{Datasets and Implement Details}
 
We evaluate PBSeg on two UAV semantic segmentation benchmarks. UAVid \cite{lyu2020uavid} contains 420 high-resolution urban images captured from low-altitude flights with 8 semantic classes (building, road, tree, low vegetation, static car, moving car, human and background clutter), split into 200 training, 70 validation and 150 test images. UDD6 \cite{chen2018large} comprises 141 images with 6 categories (facade, road, vegetation, vehicle, roof and other), divided into 90 training, 16 validation and 35 test images. All images are cropped into non-overlapping $1024 \times 1024$ patches. PBSeg is trained using the AdamW optimizer with an initial learning rate of $10^{-4}$ and cosine annealing, where the backbone learning rate is scaled by 0.1. Training proceeds for 80k iterations on UAVid and 40k iterations on UDD6 with a batch size of 16. We adopt a ResNet-50 backbone pretrained on ImageNet-1k and a transformer decoder with 6 layers, hidden dimension 256, 8 attention heads and 100 object queries. The optimization follows a standard mask-classification objective combining binary cross-entropy for class and mask logits with Dice loss, and data augmentation includes random horizontal flipping, random scaling and color jittering. All experiments are conducted on NVIDIA 3090 GPUs.

\begin{figure*}[ht]
	\centering
	\includegraphics[width=\linewidth]{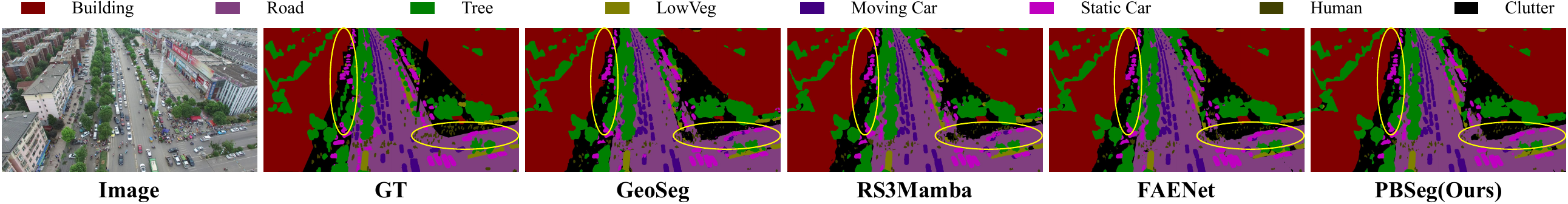}
	\caption{Visualization comparisons on the UAVid dataset. Image represents the input images, and GT represents ground-truth.
	}
	\label{f4}
\end{figure*}

\begin{figure*}[ht]
	\centering
	\includegraphics[width=\linewidth]{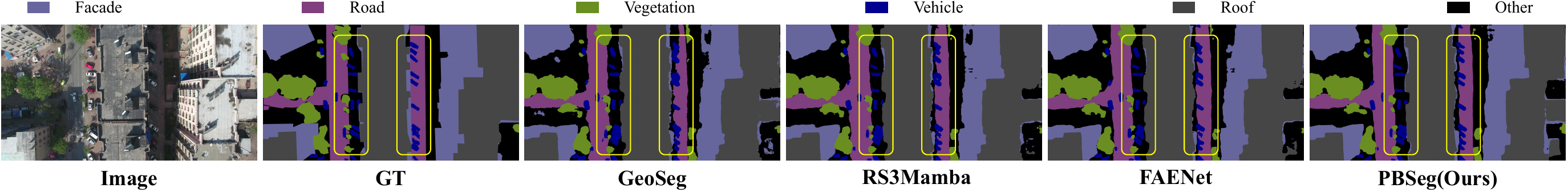}
	\caption{Visualization comparisons on the UDD6 dataset. Image represents the input images, and GT represents ground-truth.
	}
	\label{f5}
\end{figure*}

\subsection{Comparisons with SOTA Methods}
 
We compare PBSeg against recent semantic segmentation methods on both UAVid and UDD6 datasets. Quantitative results are presented in Tables \ref{tab:uavid_quant} and \ref{tab:udd6_quant}, while qualitative comparisons are shown in Figures \ref{f4} and \ref{f5}.

\textbf{Results on UAVid.} Table \ref{tab:uavid_quant} summarizes the performance of different methods on the UAVid validation set. PBSeg achieves 71.86\% mIoU, surpassing the previous best method FAENet (68.84\% mIoU) by 3.02\%. The performance gains are particularly pronounced on challenging categories. For building segmentation, PBSeg reaches 92.13\% IoU, exceeding DC-Swin by 4.49\%. On the human class, which represents small and often occluded objects in UAV imagery, PBSeg achieves 42.25\% IoU compared to FAENet's 30.98\%, an 11.27\% absolute improvement that validates the effectiveness of our prototype-based attention in capturing fine-grained details. PBSeg also excels on static car segmentation with 70.83\% IoU. Furthermore, PBSeg attains the highest mF1 score of 82.91\% and overall accuracy of 88.01\%.

\textbf{Results on UDD6.} Table \ref{tab:udd6_quant} presents results on the UDD6 test set. PBSeg achieves 80.92\% mIoU, outperforming UAV-FAENet (80.11\% mIoU) by 0.81\%. PBSeg obtains the best performance on facade (76.67\% IoU), road (73.71\% IoU), and vehicle (75.59\% IoU) classes. The 2.32\% improvement on vehicle segmentation compared to UAV-FAENet is particularly noteworthy, as vehicles constitute small objects with high variability in UAV scenes. PBSeg also records the highest overall accuracy of 88.97\%, demonstrating strong performance across diverse object scales and semantic categories in urban UAV imagery.

\textbf{Qualitative Results.} Figures \ref{f4} and \ref{f5} provide visual comparisons on representative samples from both datasets. On the UAVid example, PBSeg produces cleaner segmentation boundaries compared to competing methods, particularly on the road and tree regions where GeoSeg, RS3Mamba, and FAENet exhibit noticeable fragmentation. The highlighted circular region shows that PBSeg better preserves the integrity of small structures, whereas other methods introduce spurious predictions or miss fine details. On the UDD6 example, PBSeg demonstrates superior performance in delineating narrow road boundaries with accurate edge localization. The yellow-outlined regions reveal that GeoSeg, RS3Mamba, and FAENet produce irregular or incomplete road segmentation, while PBSeg maintains consistent predictions aligned with ground truth. These qualitative results confirm that the prototype-based attention mechanism effectively captures spatial coherence and fine-grained object boundaries in complex UAV scenes.

\subsection{Ablation Study}

To validate the effectiveness of individual components in PBSeg, we conduct ablation experiments on both UAVid and UDD6 datasets. Table \ref{tab:ablation} reports the contribution of each module.

\begin{table}[t]
	\caption{Ablation study on UAVid and UDD6 datasets}
	\label{tab:ablation}
	\centering
\resizebox{\linewidth}{!}{
	\renewcommand{\arraystretch}{1.1}
	\begin{tabular}{ccc|ccc|ccc}
		\hline
		\multirow{2}{*}{PBCA} & \multirow{2}{*}{CAM} & \multirow{2}{*}{DConv} & \multicolumn{3}{c|}{UAVid} & \multicolumn{3}{c}{UDD6} \\
		\cline{4-9}
		& & & mIoU & mF1 & OA & mIoU & mF1 & OA \\
		\hline
		\xmark & \xmark & \xmark & 66.84 & 78.52 & 85.31 & 77.15 & 86.48 & 86.92 \\
		\cmark & \xmark & \xmark & 69.23 & 80.38 & 86.54 & 78.91 & 87.56 & 87.74 \\
		\xmark & \cmark & \xmark & 68.15 & 79.47 & 85.89 & 77.82 & 86.95 & 87.28 \\
		\xmark & \xmark & \cmark & 67.96 & 79.21 & 85.72 & 77.58 & 86.81 & 87.15 \\
		\cmark & \cmark & \xmark & 70.45 & 81.28 & 87.35 & 79.74 & 88.11 & 88.42 \\
		\cmark & \xmark & \cmark & 70.08 & 80.95 & 87.08 & 79.38 & 87.89 & 88.21 \\
		\xmark & \cmark & \cmark & 69.12 & 80.15 & 86.47 & 78.65 & 87.48 & 87.81 \\
		\cmark & \cmark & \cmark & \textbf{71.86} & \textbf{82.91} & \textbf{88.01} & \textbf{80.92} & \textbf{88.56} & \textbf{88.97} \\
		\hline
	\end{tabular}
}
	
\end{table}

The baseline model (without PBCA, CAM, or DConv) achieves 66.84\% mIoU on UAVid and 77.15\% mIoU on UDD6. Incorporating PBCA alone yields significant improvements of 2.39\% mIoU on UAVid and 1.76\% mIoU on UDD6. Adding CAM or DConv individually also boosts performance, demonstrating their complementary benefits. The full PBSeg model, combining all three components, achieves the best results. This confirms that each proposed module contributes positively to the overall segmentation quality.
 
\subsection{Efficiency Analysis}

\begin{table}[t]
	\caption{Efficiency analysis on UAVid dataset}
	\label{tab:efficiency}
	\centering
	\scriptsize
	\renewcommand{\arraystretch}{1.1}
	\resizebox{\linewidth}{!}{
	\begin{tabular}{c|cccc|cc}
		\hline
		Method & FPS & GFLOPs & Latency (ms) & Para. (M) & mIoU & mF1 \\
		\hline
		ABCNet   & 102.2 & 62.9  & 9.8  & 14.0  & 66.40 & 78.13 \\
		BANet    & 37.3  & 107.2 & 26.8 & 15.44 & 66.00 & 77.86 \\
		MANet    & 75.6  & 51.7  & 13.2 & 12.0  & 66.66 & 78.50 \\
		DC-Swin  & 23.5  & 170.3 & 42.6 & 45.6  & 67.70 & 79.01 \\
		GeoSeg   & 115.6 & 46.9  & 8.7  & 11.7  & 57.18 & 70.42 \\
		CM-UNet  & 47.7  & 83.8  & 21.0 & 8.7   & 59.45 & 72.24 \\
		RS3Mamba & 17.7  & 226.0 & 56.5 & 43.32 & 68.15 & 79.51 \\
		PMamba   & 49.9  & 80.2  & 20.0 & 25.28 & 66.79 & 78.43 \\
		FAENet   & 52.0  & 76.9  & 19.2 & 19.2  & 68.84 & 79.97 \\
		\hline
		PBSeg(Ours)   & 46.6  & 85.4  & 42.9 & 35.5  & \textbf{71.86} & \textbf{82.91} \\
		\hline
	\end{tabular}
}
\end{table}

We analyze the computational efficiency of PBSeg in terms of inference speed, computational cost, and model size. Table \ref{tab:efficiency} compares FPS, GFLOPs, latency, parameter count, and accuracy metrics on the UAVid dataset.
PBSeg achieves the highest mIoU (71.86\%) and mF1 (82.91\%) among all methods. It maintains a reasonable computational cost of 85.4 GFLOPs, lower than heavy models like DC-Swin (170.3 GFLOPs) and RS3Mamba (226.0 GFLOPs). Compared to FAENet, PBSeg improves mIoU by 3.02\% with only a modest increase in GFLOPs. These results indicate that PBSeg achieves a favorable accuracy-efficiency trade-off suitable for UAV deployment scenarios.

\section{Conclusion}

This paper presents PBSeg, an efficient prototype-based segmentation framework for low-altitude UAV imagery. By introducing prototype-based cross-attention, context-aware modulation, and deformable convolutions, PBSeg achieves state-of-the-art performance on UAVid (71.86\% mIoU) and UDD6 (80.92\% mIoU) while maintaining reasonable computational efficiency. Future work will explore dynamic prototype selection and knowledge distillation to further reduce model size for real-time UAV deployment.

\bibliographystyle{IEEEbib}
\bibliography{icme2026references}

\end{document}